%% file: main.tex
\input{00_preamble}

\title{Temporal Taskification in Streaming Continual Learning: A Source of Evaluation Instability}

\author{
  Nicolae Filat \\
  Bitdefender, Romania \\
  KTH Royal Institute of Technology\\
  \texttt{nfilat@bitdefender.com}
  \And
  Elena Burceanu \\
  Bitdefender, Romania \\ 
  Politehnica University of Bucharest, Romania \\
  \texttt{eburceanu@bitdefender.com}
  \And
  Ahmed Hussain\\
  Stockholm, Sweden\\
  Network Systems Security (NSS) Group\\
  KTH Royal Institute of Technology\\
  \texttt{ahmhus@kth.se}
  \And
  Konstantinos Kalogiannis\\
  Stockholm, Sweden\\
  Network Systems Security (NSS) Group\\
  KTH Royal Institute of Technology\\
  \texttt{konkal@kth.se}
}

\preprintcopy 

\begin{document}

\maketitle

\begin{abstract}
Streaming Continual Learning (CL) typically converts a continuous stream into a sequence of discrete tasks through temporal partitioning. We argue that this temporal taskification step is not a neutral preprocessing choice, but a structural component of evaluation: different valid splits of the same stream can induce different CL regimes and therefore different benchmark conclusions. To study this effect, we introduce a taskification-level framework based on plasticity and stability profiles, a profile distance between taskifications, and Boundary-Profile Sensitivity (BPS), which diagnoses how strongly small boundary perturbations alter the induced regime before any CL model is trained. We evaluate continual finetuning, Experience Replay, Elastic Weight Consolidation, and Learning without Forgetting on network traffic forecasting with CESNET-Timeseries24, keeping the stream, model, and training budget fixed while varying only the temporal taskification. Across 9-, 30-, and 44-day splits, we observe substantial changes in forecasting error, forgetting, and backward transfer, showing that taskification alone can materially affect CL evaluation. We further find that shorter taskifications induce noisier distribution-level patterns, larger structural distances, and higher BPS, indicating greater sensitivity to boundary perturbations. These results show that benchmark conclusions in streaming CL depend not only on the learner and the data stream, but also on how that stream is taskified, motivating temporal taskification as a first-class evaluation variable.
\end{abstract}

\input{01_introduction}
\input{01_related_work}
\input{02_our_approach}
\input{03_experiments}
\input{04_limitations}
\input{05_conclusion}

\newpage
\bibliography{collas2026_conference}
\bibliographystyle{collas2026_conference}

\end{document}

%% file: 00_preamble.tex
\documentclass{article} % For LaTeX2e
\usepackage{collas2026_conference,times}
\usepackage{easyReview}

\usepackage{amsthm}

\newtheorem{theorem}{Theorem}[section]

\newtheorem{proposition}[theorem]{Proposition}

\usepackage[utf8]{inputenc} % allow utf-8 input
\usepackage[T1]{fontenc}    % use 8-bit T1 fonts
\usepackage{hyperref}       % hyperlinks
\usepackage{url}            % simple URL typesetting
\usepackage{booktabs}       % professional-quality tables
\usepackage{amsfonts}       % blackboard math symbols
\usepackage{nicefrac}       % compact symbols for 1/2, etc.
\usepackage{microtype}      % microtypography
\usepackage{xcolor}         % colors
\usepackage{multirow}
\usepackage{wrapfig,lipsum,booktabs}
\usepackage{booktabs}
\usepackage{graphicx}
\usepackage{amsmath}
\usepackage{subcaption}
\usepackage{placeins}
\usepackage{pifont}% http://ctan.org/pkg/pifont
\usepackage{acronym}

\input{00_acronyms}

\usepackage{hyperref}
\hypersetup{
    colorlinks=true,
    linkcolor=red,
    filecolor=magenta,
    urlcolor=blue,
    citecolor=purple,
    pdftitle={TEMPORAL TASKIFICATION IN STREAMING CONTINUAL LEARNING: A SOURCE OF EVALUATION INSTABILITY},
    pdfpagemode=FullScreen,
    }

%% file: 00_acronyms.tex
\acrodef{CL}{Continual Learning}
\acrodef{GCL}{General \ac{CL}}
\acrodef{ER}{Experience Replay}
\acrodef{MSE}{Mean Squared Error}
\acrodef{ISP}{Internet Service Provider}
\acrodef{BWT}{Backward transfer}
\acrodef{BPS}{Boundary-Profile Stability}
\acrodef{ML}{Machine Learning}
\acrodef{NN}{Neural Network}

%% file: 01_introduction.tex
\section{Introduction}
\acp{NN} trained on non-stationary data suffer from catastrophic forgetting, in which models overwrite previously learned knowledge in favor of newly learned information~\citep{goodfellow2015empirical}. \ac{CL} addresses this problem by accumulating knowledge over time without sacrificing past performance~\citep{chen2016lifelong}. However, solutions like weight regularization~\citep{kirkpatrick2017ewc}, dynamic architectures~\citep{serra2018overcoming}, or exemplar-free representation~\citep{gomez2024exemplar} assume that the tasks arrive as discrete units~\citep{lesort2019continual}. In practice, real-world systems operate on continuous streams where the input data are constructed rather than observed as in a dataset.

Further, several works measure \ac{CL} performance assuming prefixed tasks~\citep{chaudhry2018riemannian, kemker2018measuring, de2021continual}, while investigations on the effect of task ordering~\citep{yoon2020scalable, buzzega2020dark} also assume a given temporal partition of the tasks. Existing research in \ac{CL} for time series typically treats task boundaries as a given property of the experimental setup~\citep{besnard2024continual, wang2021inclstm, gupta2021continual, du2021adarnn, kim2022reversible}. This is also observed in works discussing concept drift~\citep{gama2014survey, lu2019learning}. By relying on fixed time windows, these works overlook the potential bias introduced by the windowing step itself, providing evidence that the construction of evaluation tasks has received insufficient attention.

Unlike \ac{CL} benchmarks built from static datasets such as CIFAR~\citep{cifar} and ImageNet~\citep{deng2009imagenet}, network telemetry evolves as a genuine stream, with temporally structured variation, recurring patterns, and natural non-stationarity. This makes the network domain a suitable setting for studying whether temporal task construction captures the underlying learning problem or instead introduces additional experimental bias. 

In this work, we explicitly question the task-boundaries assumption by investigating how segmenting a continuous stream into tasks can alter performance metrics. Our \textbf{main contributions} are as follows:

\begin{itemize}
    \item \textbf{Temporal taskification matters.} We formalize temporal taskification as a structural component of evaluation in streaming \ac{CL}, and show that different valid splits of the same stream can induce different \ac{CL} regimes.
    
    \item \textbf{Taskifications can be diagnosed efficiently before training.} We develop a taskification-level framework based on plasticity and stability profiles, profile distance, and \ac{BPS}, which enables comparing temporal splits and assessing their robustness before training any \ac{CL} model.
    
    \item \textbf{Structural variation predicts evaluation variability.} On the CESNET-Timeseries24 dataset, we show that temporal taskifications with different induced distributional structures also differ in profile distance and robustness, and that these differences translate into materially different \ac{CL} conclusions, including changes in forecasting error, forgetting, and backward transfer.
\end{itemize}

\paragraph{Paper Structure.} Sec.~\ref{sec:related_work} situates our work within the \ac{CL}, time-series forecasting, and benchmark robustness literature. Sec.~\ref{sec:problem_formulation} formalizes temporal taskification, and Sec.~\ref{sec:approach} introduces plasticity and stability profiles, defines the profile distance, and presents \ac{BPS} as a pre-training diagnostic of taskification robustness. Sec.~\ref{sec:experiments} describes the experimental setup and reports results across the 9-, 30-, and 44-day taskifications for four \ac{CL} methods. Sec.~\ref{sec:limitations} discusses limitations, and Sec.~\ref{sec:conclusions} concludes the paper and outlines directions for future work.

%% file: 01_related_work.tex
\section{Related Work}
\label{sec:related_work}

\paragraph{Relation to Continual Learning Evaluation.} Prior work focuses on measuring \ac{CL} performance once tasks are fixed.~\citep{chaudhry2018riemannian} formalize forgetting and intransigence as complementary evaluation metrics, while~\citep{kemker2018measuring} benchmark catastrophic forgetting across multiple mitigation strategies, and~\citep{delange2022continual} systematize the stability--plasticity trade-off under fixed evaluation protocols.~\citep{hsu2018reevaluating} further show that even scenario categorization significantly changes benchmark conclusions. These studies assume that task structure is predetermined; our work goes a step earlier and demonstrates that \emph{temporal taskification} is itself a structural evaluation variable whose choice alters the measured forgetting, transfer, and stability--plasticity profiles.

\paragraph{Relation to Task Order and Curriculum Effects in Continual Learning.}~\citep{bengio2009curriculum} established that the ordering of training examples shapes both convergence speed and generalization, a principle that naturally extends to \ac{CL} where tasks arrive sequentially. Within \ac{CL} specifically,~\citep{yoon2020scalable} introduce order-robustness metrics and propose additive parameter decomposition to mitigate sensitivity to permutations of a fixed task sequence, while~\citep{buzzega2020dark} blur task boundaries in their \ac{GCL} setting and show that rehearsal with distillation provides strong baselines even as boundaries soften. These works demonstrate that task order matters given a fixed set of tasks; our contribution operates at a logically prior level by showing that, in streaming settings, the task sequence itself is an artifact of how the continuous stream is partitioned; different valid temporal splits produce different orderings, different numbers of tasks, and ultimately different benchmark conclusions.

\paragraph{Relation to Online and Streaming Continual Learning.} Online and streaming \ac{CL} methods process data without revisiting previous samples:~\citep{aljundi2019online} select replay samples via maximal retrieval interference,~\citep{caccia2022new} diagnose how abrupt representation changes degrade online learners, and~\citep{mai2022online} provide a systematic empirical comparison of online \ac{CL} methods under standardized evaluation protocols. More recently,~\citep{ghunaim2023realtime} argue that evaluation itself must be rethought for the online setting by proposing real-time metrics that assess performance during stream processing rather than only at task boundaries. Despite these advances, evaluation in all of these works still depends (implicitly or explicitly) on a temporal split that defines when one phase of the stream ends, and another begins. Our work makes this dependence explicit by analyzing how the choice of temporal partition changes the induced \ac{CL} regime and the conclusions one draws about method performance.

\paragraph{Relation to Task-Free and Task-Agnostic Continual Learning.} Task-free and task-agnostic methods aim to remove dependence on explicit task labels during training:~\citep{lee2020neural} use a neural Dirichlet process mixture to automatically discover task structure without boundaries, and~\citep{vandeven2022three} formalize the taxonomy of incremental learning scenarios (task-, domain-, and class-incremental) showing that the assumed task structure fundamentally determines which methods succeed.~\citep{prabhu2023computationally} further demonstrate that under realistic computational budgets, the gap between task-aware and task-free methods narrows substantially, questioning whether task labels provide a genuine signal or merely evaluation scaffolding. While these works eliminate task identity during training, evaluation typically reintroduces task structure, leading to per-task metrics such as forgetting or backward transfer. Our contribution focuses specifically on this evaluation-side dependence: 
the choice of temporal taskification for evaluation is not neutral and induces different stability--plasticity profiles and benchmark rankings.

\paragraph{Relation to Benchmark Robustness and Evaluation Sensitivity.} A growing body of work exposes fragilities in how \ac{ML} progress is measured.~\citep{recht2019imagenet} demonstrate that ImageNet accuracy drops significantly on a carefully re-collected test set, revealing benchmark overfitting, while~\citep{torralba2011unbiased} show that dataset-specific biases allow classifiers to identify which dataset an image comes from rather than learning generalizable features. At the methodological level,~\citep{bouthillier2021accounting} quantify how sources of variance (e.g., random seeds, hyperparameters, data splits) can reverse method rankings, and~\citep{dehghani2021benchmark} formalize the benchmark lottery, showing that which benchmarks are selected determines which methods appear state-of-the-art. Our work identifies a previously uncharacterized source of benchmark instability specific to streaming \ac{CL}: temporal taskification. We show that \ac{BPS} can quantify this instability before any model is trained, complementing the general benchmark-robustness literature with a domain-specific structural analysis.

%% file: 02_our_approach.tex
\section{Problem Formulation}
\label{sec:problem_formulation}
We start from a single temporally ordered stream and study how different temporal partitions of that same stream affect \ac{CL} evaluation. Formally, a temporal taskification is an ordered partition
\begin{equation}
\tau = (t_0,\dots,t_K),
\qquad
0=t_0<t_1<\dots<t_K=T,
\end{equation}
which segments the stream into task intervals $I_k^\tau := [t_{k-1},t_k)$ for $k=1,\dots,K$. Each interval induces a task-level distribution $P_k^\tau$, yielding a sequence of distributions
$(P_1^\tau,\dots,P_K^\tau)$.

Different valid taskifications of the same stream may therefore induce different numbers of tasks, different adjacent transitions, and different longer-range recurrence structure, even though the underlying data source is unchanged. Our goal is to compare such taskifications at the structural level, before training any \ac{CL} method.

In particular, we seek to answer two questions: (i) when do two taskifications induce meaningfully different \ac{CL} regimes, and (ii) when is a chosen taskification structurally robust or structurally fragile under small perturbations of its boundaries? To this end, we represent each taskification through task-count-invariant plasticity and stability profiles, define a profile distance between taskifications, and measure the structural sensitivity of a chosen split through local boundary perturbations.

\section{Proposed Approach}
\label{sec:approach}

\subsection{Plasticity and Stability Profiles}
\label{subsec:profiles}
Different temporal taskifications may yield different numbers of tasks, making task-indexed comparisons inconvenient and often ill-defined. To obtain a representation that is comparable across split families and computationally cheap, we summarize each taskification using two empirical profile distributions: a plasticity profile and a stability profile. Let a taskification $\tau$ induce task-level distributions $P_1^\tau,\dots,P_{K_\tau}^\tau$.

\paragraph{Plasticity profile.}
We construct the plasticity profile of $\tau$ from the set of discrepancies between consecutive tasks, namely $d(P_k^\tau,P_{k+1}^\tau)$ for $k=1,\dots,K_\tau-1$, where $d(\cdot,\cdot)$ is a distance or divergence between task distributions. Rather than keeping these values as a task-indexed sequence, we treat them as samples from an empirical distribution, denoted by $\Pi_{\mathrm{pl}}^\tau$. Intuitively, $\Pi_{\mathrm{pl}}^\tau$ captures how frequently the chosen taskification induces mild, moderate, or abrupt transitions between adjacent tasks.

\paragraph{Stability profile.}
We construct the stability profile from longer-range relations in the induced task sequence. Specifically, we consider the distance or divergence $d(P_i^\tau,P_j^\tau)$ between non-adjacent tasks, for pairs $(i,j)$ satisfying $j-i \ge \ell_{\min}$, where $\ell_{\min}$ excludes immediate neighbors so that the profile reflects longer-range recurrence rather than local transitions. We again treat these values as samples from an empirical distribution, denoted by $\Pi_{\mathrm{st}}^\tau$. Intuitively, $\Pi_{\mathrm{st}}^\tau$ reflects the extent to which previously observed patterns differ from those that appear in the future.

This representation has two advantages. First, it is invariant to the number of induced tasks, since each taskification is represented by distributions of transition distances rather than by fixed-length vectors. Second, it is cheap to compute, since it only requires computing task-level distributions and pairwise distances or divergences, without training any \ac{CL} model.

\subsection{Profile Distance}
\label{subsec:profile_distance}

The profile representation above provides a structural, task-count-invariant description of a temporal taskification. We compare two taskifications through the discrepancy between their induced plasticity and stability profiles.

Let $D_{\mathrm{pl}}(\tau,\sigma)$ denote a distribution distance between the plasticity profiles $\Pi_{\mathrm{pl}}^\tau$ and $\Pi_{\mathrm{pl}}^\sigma$, and let $D_{\mathrm{st}}(\tau,\sigma)$ denote the corresponding distance between the stability profiles $\Pi_{\mathrm{st}}^\tau$ and $\Pi_{\mathrm{st}}^\sigma$. We then define the overall taskification profile distance as
\begin{equation}
D_{\mathrm{prof}}(\tau,\sigma)
:=
\left[
\alpha D_{\mathrm{pl}}(\tau,\sigma)^2
+
\beta D_{\mathrm{st}}(\tau,\sigma)^2
\right]^{1/2},
\label{eq:dprof}
\end{equation}
where $\alpha,\beta > 0$ are scaling coefficients.

Intuitively, $D_{\mathrm{prof}}(\tau,\sigma)$ is small when two taskifications induce similar patterns of adjacent-task change and similar patterns of longer-range recurrence, and large when they induce substantially different \ac{CL} regimes. Importantly, this distance depends only on the induced task distributions and not on the number of tasks themselves. It therefore provides a model-free structural quantity that can be computed before training and used to compare arbitrary temporal splits.

\subsection{Structural Robustness of a Taskification}
\label{subsec:structural_robustness}

A temporal taskification should not be judged only by the benchmark numbers it produces. It should also be judged by how sensitive its induced \ac{CL} regime is to small changes in boundary placement. If minor perturbations of the split produce substantially different plasticity and stability profiles, then the resulting benchmark is structurally fragile: its conclusions may depend more on arbitrary boundary placement than on stable properties of the underlying stream.

\paragraph{\acf{BPS}.} To formalize this, let $\tau$ be a taskification and let $\mathcal{N}_{\delta}^{\mathrm{bdry}}(\tau)$ denote the set of valid taskifications obtained by perturbing each internal boundary of $\tau$ by at most $\delta$ in time. We measure the structural sensitivity of $\tau$ through the mean of the profile distance over this neighborhood. Concretely, $\mathrm{BPS}(\tau;\delta)$ is the mean of $D_{\mathrm{prof}}(\tau,\sigma)$ over perturbed taskifications $\sigma \in \mathcal{N}_{\delta}^{\mathrm{bdry}}(\tau)$. Low values indicate that even relatively unfavorable perturbations leave the induced plasticity and stability profiles nearly unchanged. High values indicate structural fragility: small perturbations can move the benchmark into a materially different \ac{CL} regime.

This notion is useful because it does not require training any learner. It measures whether the chosen split is a stable substrate for evaluation before any downstream \ac{CL} metric is computed. In this sense, profile sensitivity is not a substitute for benchmark performance, but a prior diagnostic of whether benchmark conclusions are likely to be robust or arbitrary with respect to boundary placement.

\begin{figure}[t]
    \centering
    \begin{subfigure}[t]{0.32\textwidth}
        \centering
        \includegraphics[width=\linewidth]{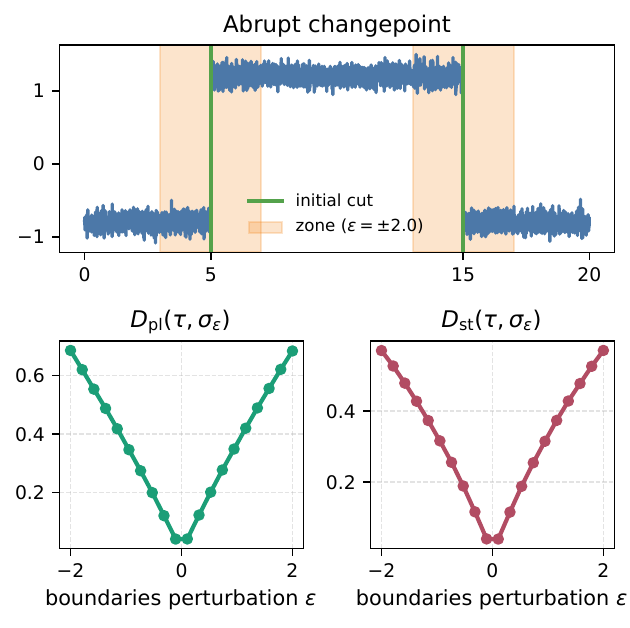}
        \caption{Abrupt changepoint: a small boundary perturbation that crosses the changepoint sharply alters adjacent-task discrepancy, producing a large change in the plasticity profile.}
        \label{fig:fragile_changepoint}
    \end{subfigure}
    \hfill
    \begin{subfigure}[t]{0.32\textwidth}
        \centering
        \includegraphics[width=\linewidth]{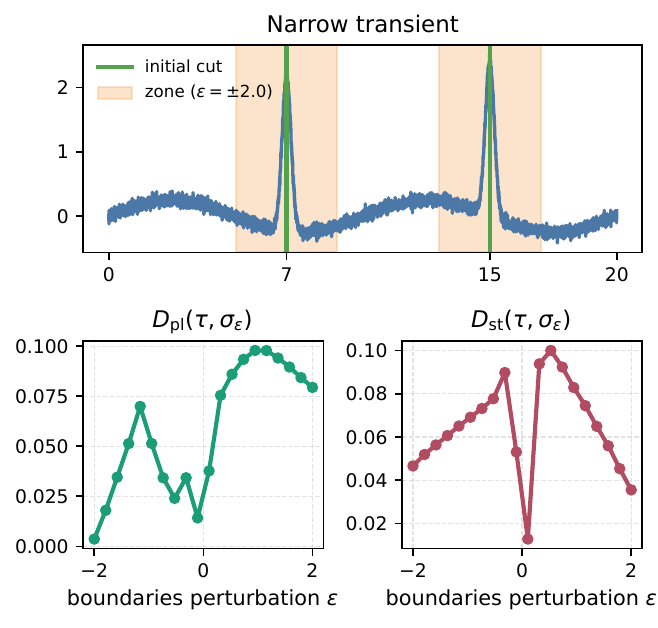}
        \caption{Narrow transient: a short high-intensity event is reassigned across neighboring tasks under a small perturbation, yielding a disproportionate change in the plasticity profile.}
        \label{fig:fragile_transient}
    \end{subfigure}
    \hfill
    \begin{subfigure}[t]{0.32\textwidth}
        \centering
        \includegraphics[width=\linewidth]{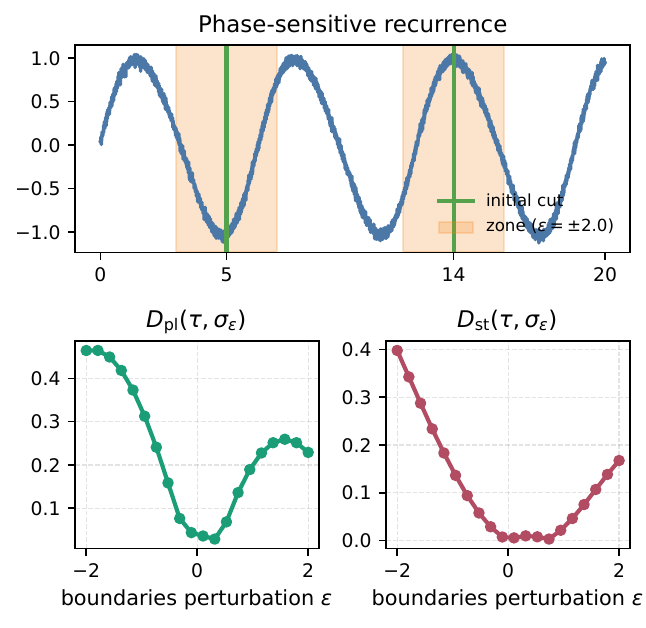}
        \caption{Phase-sensitive recurrence: a small phase shift in the boundaries changes which tasks appear recurrent, substantially perturbing the stability profile.}
        \label{fig:fragile_recurrence}
    \end{subfigure}
    \caption{Illustrative examples of structurally fragile taskifications. Small boundary perturbations can induce large profile distance through different mechanisms: (a) abrupt local transitions, (b) localized transient events, or (c) misaligned long-range recurrence.}
    \label{fig:fragile_taskifications}
\end{figure}

\paragraph{Robust vs. Fragile Taskifications.} We distinguish between two broad regimes: robust and fragile segmentations. In robust taskifications, the underlying stream varies smoothly around the chosen boundaries, so small temporal perturbations induce only small changes in the task-level distributions and, consequently, only small changes in the induced plasticity and stability profiles. In contrast, for fragile taskifications, a boundary lies near a structurally sensitive region of the stream, such as an abrupt changepoint, a narrow transient, or a phase-sensitive recurrent pattern. In such cases, even a small perturbation can alter the induced adjacent-task discrepancies or long-range recurrence structure, leading to a large profile distance and high profile sensitivity. This distinction can be illustrated with simple synthetic constructions, as shown in Fig.~\ref{fig:fragile_taskifications}.

\paragraph{Case Study 1: Abrupt changepoint.}
Consider a stream with a single distribution shift at times $t^\star_1$ and $t^\star_2$,
\begin{equation}
x_t \sim
\begin{cases}
\mathcal{N}(\mu_1,\sigma^2) & \text{if } t < t^\star_1\\
\mathcal{N}(\mu_2,\sigma^2) & \text{if } t \in [t^\star_1, t^\star_2] \\
\mathcal{N}(\mu_1,\sigma^2) & \text{if } t > t^\star_2
\end{cases}
\qquad \textrm{with }  \mu_1 \neq \mu_2.
\label{eq:abrupt_changepoint}
\end{equation}
If two nearby taskifications place boundaries close to $t^\star_1$ and $t^\star_2$, then even a small perturbation can sharply alter the adjacent-task discrepancy, producing a large change in the plasticity profile; see Fig.~\ref{fig:fragile_changepoint}.

\paragraph{Case Study 2: Narrow transient.}
Consider a smooth background process $g(t)$ with two high-intensity events at $t^\star_1$ and $t^\star_2$,
\begin{equation}
x_t = g(t) + A \exp\!\left(-\frac{(t-t^\star_1)^2}{2\eta^2}\right) + A \exp\!\left(-\frac{(t-t^\star_2)^2}{2\eta^2}\right) + \xi_t,
\label{eq:narrow_transient}
\end{equation}
with small $\eta$. If a task boundary lies near $t^\star_1$ and $t^\star_2$, a small perturbation can change whether the transient peaks are concentrated in one task or split across two tasks, yielding a disproportionate change in the plasticity profile; see Fig.~\ref{fig:fragile_transient}.

\paragraph{Case Study 3: Phase-sensitive recurrence.}
Consider a recurrent stream such as
\begin{equation}
x_t = \sin(\omega t) + \xi_t.
\label{eq:periodic_stream}
\end{equation}
If two nearby taskifications differ by a small phase shift in their boundaries, then the induced tasks may no longer align with the same phase of the cycle, substantially perturbing the long-range recurrence structure and therefore the stability profile; see Fig.~\ref{fig:fragile_recurrence}.

These examples show that some taskifications are intrinsically fragile: small boundary perturbations can already induce materially different \ac{CL} regimes. We therefore use boundary profile sensitivity as a taskification-level diagnostic, with low $\mathrm{BPS}$ indicating robustness and high $\mathrm{BPS}$ indicating split-dependent fragility.

\begin{proposition}[Informal]
There exist non-stationary streams and valid temporal taskifications such that arbitrarily small boundary perturbations induce a non-negligible profile distance and, therefore, high structural sensitivity.
\end{proposition}

This result is purely structural: it isolates a source of benchmark variability that exists before any model is trained.

%% file: 03_experiments.tex
\section{Experiments}
\label{sec:experiments}

We organize the experiments to validate the main components of our framework and their empirical consequences. We first describe the common experimental setup (Sec.~\ref{subsec:experimental_setup}). Then we examine the induced task distributions to understand how different taskifications reshape the underlying regime (Sec.~\ref{subsec:distribution_level_differences}). Next, we test whether changing only the temporal taskifications alters \ac{CL} conclusions (Sec.~\ref{subsec:taskification_sensitivity}). We then quantify structural differences between taskifications through the proposed profile distance (Sec.~\ref{subsec:profile_distance_experiments}). Finally, we evaluate boundary-profile sensitivity to identify which taskifications are more or less robust under small perturbations of their boundaries (Sec.~\ref{subsec:bps_experiments}).

\begin{figure}[t]
    \centering
    \includegraphics[width=0.98\linewidth]{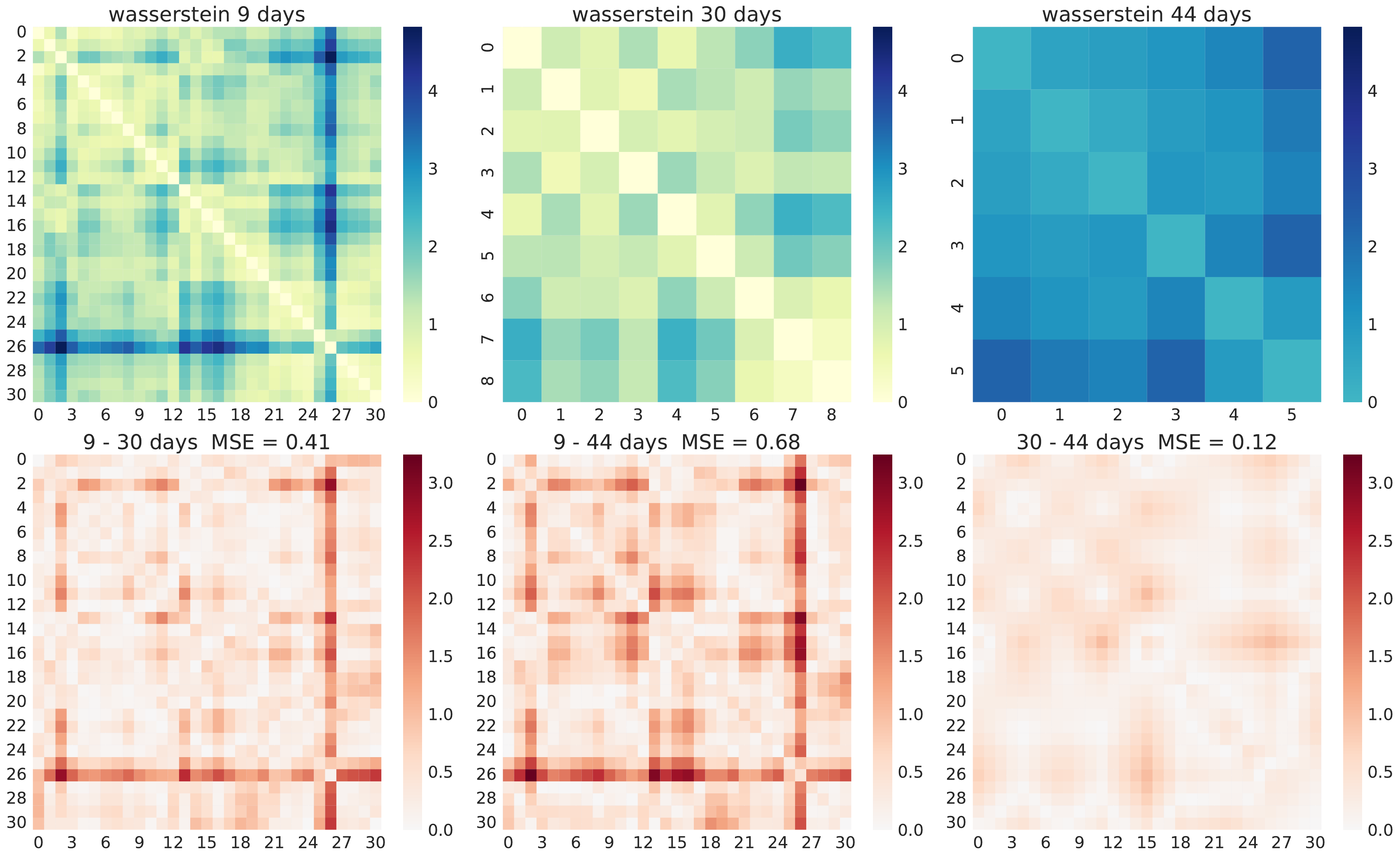}
    \caption{Top row: pairwise Wasserstein distances between induced tasks for the 9-day, 30-day, and 44-day taskifications. The 9-day split exhibits noisier and less regular transitions than the longer-window splits. Bottom row: MSE between pairs of upsampled task-to-task distance matrices and their corresponding absolute difference heatmaps. The 30-day and 44-day taskifications induce more similar structural patterns than either does relative to the 9-day split.}
    \label{fig:js}
\end{figure}

\subsection{Experimental Setup}
\label{subsec:experimental_setup}

\paragraph{Dataset.}
We use the CESNET-Timeseries24 dataset~\citep{cesnet24}, which contains 40 weeks of network traffic collected from a Czech university \ac{ISP}. From this data, we selected the $100$ highest-density IP addresses for the $10$ minutes aggregation interval.

\paragraph{Forecasting task and temporal segmentation.}
Our setting is multivariate time-series forecasting. For each selected IP address, the objective is to predict the \texttt{avg\_duration} feature from past observations of all available variables. We treat the data as a single temporally ordered stream and construct continual-learning tasks by partitioning it into fixed-length temporal windows. Each task, therefore, corresponds to a contiguous time interval and contains the $100$ time series associated with the selected IPs over that interval.

In the experiments below, we consider three temporal taskifications with window lengths of $9$, $30$, and $44$ days. All three lengths are congruent to $2 \pmod 7$, which preserves weekday alignment across splits. Within each task, $80\%$ of the samples are used for training, $10\%$ for validation, and $10\%$ for testing.

\paragraph{Model and preprocessing.}
For all taskifications, we use the same forecasting model, a time-series-based Transformer, with two days of context as input and the next 10-minute timestep as the prediction target. Each input contains $12$ features, which are linearly projected into a $32$-dimensional latent space and processed by a stack of $8$ Transformer blocks with $4$ attention heads each. To stabilize training across heterogeneous feature scales, we apply per-sample normalization by standardizing each feature over the temporal dimension, following~\citep{chronosv2}. We additionally normalize the target feature within each task by dividing it by the maximum value observed in that task's training split, ensuring the loss remains comparable across IPs with different scales. Since this normalization leads to small absolute errors, we report all \ac{MSE} values multiplied by $10^3$ for readability.

\paragraph{Training and evaluation protocol.}
The model is trained with AdamW using a learning rate of $10^{-4}$ and a batch size of $256$. We reduce the learning rate by a factor of $0.1$ when the validation loss does not improve for $5$ consecutive epochs, and we apply early stopping with a patience of $10$ epochs.

Across all experiments, we compare continual finetuning, Experience Replay~\citep{er}, Elastic Weight Consolidation~\citep{ewc}, and Learning without Forgetting~\citep{lwf}. Performance is measured with average \ac{MSE}, Backward Transfer, and Forgetting. Since the latter two are traditionally defined for classification accuracy, we adapt them to the regression setting by defining $M_{i,j}$ as the \ac{MSE} on task $j$ after training on task $i$.

For a total of $T$ tasks, forgetting is computed as $f_j = M_{T,j} - \min_{k \in \{j,\dots,T-1\}} M_{k,j}$ and averaged over past tasks as
$\mathrm{Forgetting} = \frac{1}{T-1}\sum_{j=1}^{T-1} f_j$. Similarly, Backward Transfer is defined as $\mathrm{BWT} = \frac{1}{T-1}\sum_{j=1}^{T-1} \left(M_{j,j} - M_{T,j}\right)$.

These adaptations ensure that while the mathematical operations are inverted to suit a minimization objective, the underlying concepts being measured remain identical to those established in the foundational survey by~\citet{wang2024comprehensive}.

\paragraph{\ac{BPS} implementation.}
The proposed \ac{BPS} metric is implemented by employing the first-order Wasserstein distance as the discrepancy measure $d(\cdot, \cdot)$. We define the boundary neighborhood $\mathcal{N}_{\delta}^{\mathrm{bdry}}$ via a random perturbation of each boundary by up to $\delta = \pm 1$ day. To balance the contributions of plasticity and stability, the coefficients $\alpha$ and $\beta$ are assigned equal weight ($\alpha = \beta = 0.5$).

\subsection{Distribution-Level Differences Across Taskifications} 
\label{subsec:distribution_level_differences}

We first examine whether different temporal taskifications induce visibly different task-level distribution patterns. This provides a structural view of the benchmark before analyzing downstream \ac{CL} metrics.

To quantify distribution differences across temporal scales, we compute pairwise Wasserstein distances between induced tasks. Fig.~\ref{fig:js} shows the resulting task-to-task distance matrices for the 9-day, 30-day, and 44-day taskifications. Since these matrices have different sizes, we upsample them to a common resolution and compute the mean squared error between each pair, enabling a direct comparison across taskifications.

The resulting patterns differ markedly across splits. The 30-day and 44-day taskifications exhibit smoother and more gradually evolving structures, suggesting that longer windows capture more stable temporal regimes. In contrast, the 9-day split shows higher-frequency and more irregular transitions, indicating a noisier task sequence. The pairwise matrix comparisons are consistent with this observation: taskifications that differ more strongly in window length also induce larger matrix-level MSE, while the 30-day and 44-day splits remain more similar to each other than either is to the 9-day split.

These distribution-level differences provide a structural explanation and an intuition for the taskification sensitivity observed later in Tab.~\ref{tab:taskification_results}. Shorter windows appear to fragment the stream into less stable tasks, whereas the 30-day and 44-day taskifications provide a more consistent signal for optimization. This interpretation is also consistent with the boundary-profile sensitivity results discussed in Sec.~\ref{subsec:bps_experiments}.

\begin{table*}[t]
\centering
\small
\caption{Effect of temporal taskification on \ac{CL} performance. Each window size defines a different taskification of the same stream. Lower is better for average mean squared error and Forgetting, while higher is better for Backward Transfer. The last three columns report the sample standard deviation of each metric across the 9-, 30-, and 44-day taskifications for a fixed algorithm.}
\label{tab:taskification_results}
\begin{tabular}{l|ccc|ccc|ccc|ccc}
\toprule
\multirow{2}{*}{\textbf{Algorithm}} 
& \multicolumn{3}{c|}{\textbf{Average MSE} $\downarrow$}
& \multicolumn{3}{c|}{\textbf{BWT} $\uparrow$}
& \multicolumn{3}{c|}{\textbf{Forgetting} $\downarrow$}
& \multicolumn{3}{c}{\textbf{Std. across taskifications}} \\
\cmidrule(lr){2-4} \cmidrule(lr){5-7} \cmidrule(lr){8-10} \cmidrule(lr){11-13}
& 9d & 30d & 44d & 9d & 30d & 44d & 9d & 30d & 44d 
& MSE Std. & BWT Std. & Fgt Std. \\
\midrule
ER       
& 13.04 & 10.88 & 30.88
& 1.67 & 0.45 & -1.95
& 0.20 & -0.01 & 2.15
& 10.98 & 1.84 & 1.19 \\

EWC      
& 13.43 & 10.77 & 26.95
& 0.60 & 0.11 & 0.75
& -0.03 & -0.01 & -0.27
& 8.68 & 0.33 & 0.14 \\

Finetune 
& 12.58 & 10.58 & 30.38
& 0.36 & 0.10 & -5.65
& 0.08 & 0.11 & 5.71
& 10.90 & 3.40 & 3.24 \\

LwF      
& 12.57 & 10.52 & 27.04
& 0.89 & 0.21 & 0.62
& 0.07 & 0.10 & -0.54
& 9.00 & 0.34 & 0.36 \\
\bottomrule
\end{tabular}
\end{table*}

\subsection{Taskification Sensitivity}
\label{subsec:taskification_sensitivity}
We now test the central empirical question of the paper: whether changing only the temporal taskification can alter \ac{CL} conclusions while keeping the stream, model, and training budget fixed. Tab.~\ref{tab:taskification_results} shows that the answer is clearly yes. Across all four methods, the measured performance varies substantially across the 9-day, 30-day, and 44-day taskifications. 

A consistent pattern emerges for average MSE: the 30-day split yields the lowest error across all methods, whereas the 44-day split results in the strongest degradation. This indicates that the temporal partition alone can change the benchmark's predictive difficulty.  The last three columns of Tab.~\ref{tab:taskification_results} summarize this effect through the sample standard deviation of each metric across the three taskifications for a fixed method. These values provide a compact measure of taskification sensitivity and reinforce the same conclusion: benchmark outcomes in \ac{CL} depend not only on the learner and the stream, but also on how the stream is partitioned into tasks.

\subsection{Profile Distance Between Taskifications}
\label{subsec:profile_distance_experiments}

We next ask whether the temporal taskifications considered in this work induce meaningfully different \ac{CL} regimes in the structural sense captured by our framework. To quantify this difference, we compute the profile distance \(D_{\mathrm{prof}}\) introduced in Eq.~\ref{eq:dprof}, which compares taskifications through both their plasticity and stability profiles.

Tab.~\ref{tab:d_prof} shows that the three taskifications are indeed structurally distinct. As expected, the profile distance increases as the difference in task duration becomes larger. In particular, the 9-day and 44-day taskifications are the most dissimilar, while the 30-day and 44-day taskifications are the closest pair. This ordering is consistent with the distribution-level observations in Fig.~\ref{fig:js}, where the 30-day and 44-day splits exhibit more similar task-to-task structure than either does relative to the 9-day split.

These results support the interpretation that the three temporal splits do not merely represent different granularities of the same benchmark, but induce measurably different structural regimes from the perspective of \ac{CL}.

\begin{table}[t]
\centering
\caption{\(D_{\mathrm{prof}}\) mean \(\pm\) standard deviation across 100 IPs for different taskification pairs.}
\begin{tabular}{lccc}
\toprule
 & 9 days & 30 days & 44 days \\
\midrule
9 days  & --              & 0.85 $\pm$ \small{1.1} & 1.19 $\pm$ \small{1.6} \\
30 days & 0.85 $\pm$ \small{1.2} & --              & 0.75 $\pm$ \small{0.9} \\
44 days & 1.19 $\pm$ \small{1.6} & 0.75 $\pm$ \small{0.9} & --              \\
\bottomrule
\end{tabular}
\label{tab:d_prof}
\end{table}

\subsection{Boundary-Profile Sensitivity (BPS)} 
\label{subsec:bps_experiments}

We evaluate which temporal taskifications are structurally more robust under small boundary perturbations. To this end, we use the \acf{BPS} introduced in Sec.~\ref{sec:approach}. Recall that high \ac{BPS} means that small changes in task boundaries induce large changes in the plasticity and stability profiles, indicating a structurally fragile taskification.

Tab.~\ref{tab:BPS_standard} shows a clear trend across the three window lengths. As the task duration increases, the mean plasticity, stability, and \ac{BPS} values decrease. In particular, the 9-day split has the highest \ac{BPS}, followed by the 30-day split, while the 44-day split is the least sensitive. This indicates that shorter taskifications are structurally less stable, in the sense that small boundary perturbations induce larger changes in the underlying \ac{CL} regime.

We also test whether this pattern depends strongly on the exact starting point of the segmentation. To do so, we repeat the same analysis after shifting all taskifications by two days. The results in Tab.~\ref{tab:BPS_plus_2_days} remain consistent with the original ordering: the 9-day split is still the most sensitive, the 44-day split remains the most stable, and the 30-day split stays in between. This confirms that the observed robustness pattern is not an artifact of a single alignment, but persists under a nontrivial change in boundary placement.

Taken together, these results provide direct empirical support for the robustness notion introduced in our framework. They show that \ac{BPS} captures a meaningful structural property of temporal taskifications and that shorter windows tend to produce more fragile segmentations than longer ones.

\begin{table}[t]
\centering
\caption{Mean \(\pm\) standard deviation of plasticity, stability, and \ac{BPS} across 100 IPs for each taskification. Bold indicates the highest mean in each row, and blue indicates the second highest.}

\begin{tabular}{lccc}
\toprule
 & 9 days & 30 days & 44 days \\
\midrule

Plasticity 
& \textbf{0.15} $\pm$ \small{0.16} 
& \textcolor{blue}{0.08} $\pm$ \small{0.09} 
& 0.06 $\pm$ \small{0.08} \\

Stability 
& \textbf{0.08} $\pm$ \small{0.10}
& \textcolor{blue}{0.07} $\pm$ \small{0.07} 
& 0.05 $\pm$ \small{0.07} \\

\ac{BPS} 
& \textbf{0.12} $\pm$ \small{0.13} 
& \textcolor{blue}{0.08} $\pm$ \small{0.08} 
& 0.06 $\pm$ \small{0.07} \\
\bottomrule
\end{tabular}
\label{tab:BPS_standard}
\end{table}

\begin{table}[ht!]
\centering
\caption{Mean \(\pm\) standard deviation of plasticity, stability, and \ac{BPS} across 100 IPs after shifting all task boundaries by 2 days. Bold indicates the highest mean in each row, and blue indicates the second highest.}
\begin{tabular}{lccc}
\toprule
 & 9 days + $\Delta$2 days & 30 days + $\Delta$2 days  & 44 days + $\Delta$2 days  \\
\midrule

Plasticity 
& \textbf{0.15} $\pm$ \small{0.17} 
& \textcolor{blue}{0.11} $\pm$ \small{0.12} 
& 0.06 $\pm$ \small{0.07} \\

Stability 
& \textcolor{blue}{0.08} $\pm$ \small{0.09} 
& \textbf{0.09} $\pm$ \small{0.09} 
& 0.05 $\pm$ \small{0.06} \\

\ac{BPS}
& \textbf{0.12} $\pm$ \small{0.13} 
& \textcolor{blue}{0.10} $\pm$ \small{0.10} 
& 0.05 $\pm$ \small{0.06} \\
\bottomrule
\end{tabular}
\label{tab:BPS_plus_2_days}
\end{table}

%% file: 04_limitations.tex
\section{Limitations}
\label{sec:limitations}

\paragraph{Scope of the empirical domain.}
Our empirical evaluation is restricted to a single application domain, namely, network traffic forecasting on CESNET-Timeseries24. This setting is well-suited for studying temporally evolving streams, but it remains possible that the magnitude and form of taskification sensitivity differ in other streaming domains with different temporal structure, recurrence patterns, or noise characteristics.

\paragraph{Coverage of \ac{CL} methods.}
We evaluate four \ac{CL} strategies: continual finetuning, Experience Replay, Elastic Weight Consolidation, and Learning without Forgetting. While these methods cover several standard \ac{CL} paradigms, they do not exhaust the broader space of online, task-free, rehearsal-free, or architecture-based approaches. As a result, the extent to which taskification sensitivity appears may vary across method families not considered here.

\paragraph{Family of taskifications considered.}
Our experiments focus on fixed-length temporal taskifications and local boundary perturbations. This keeps the comparison controlled: changes in plasticity and stability profiles can be traced directly to differences in window length and boundary placement. More adaptive segmentation strategies may alter these profiles in more complex ways by jointly changing task count, transition strength, and recurrence structure. We leave such taskification families to future work.

\paragraph{Diagnostic rather than prescriptive scope.}
The contribution of this work is primarily diagnostic. We provide a framework for characterizing, comparing, and stress-testing temporal taskifications before training, but we do not yet propose an automatic procedure for selecting an optimal taskification, nor a \ac{CL} method robust to taskification variability. These directions remain open for future work.

%% file: 05_conclusion.tex
\section{Conclusion}
\label{sec:conclusions}

We studied temporal taskification as a source of evaluation variability in streaming \acf{CL}. Our main finding is that partitioning a continuous stream into temporal tasks is not a neutral preprocessing step, but a structural component of the benchmark itself. Different valid taskifications of the same stream can yield different task distributions, distinct plasticity and stability profiles, and, ultimately, distinct conclusions about \ac{CL}.

To make this dependence explicit, we introduced a taskification-level framework based on plasticity and stability profiles, a profile distance between taskifications, and \ac{BPS}. These quantities provide a cheap, pre-training way to compare temporal splits and assess how robust a chosen taskification is under small boundary perturbations.

Our experiments support this view at multiple levels. On CESNET-Timeseries24, changing only the temporal taskification produces substantial differences in forecasting error, forgetting, and backward transfer across all evaluated methods. These differences are also reflected structurally: shorter taskifications induce noisier distribution-level patterns, larger sensitivity to boundary perturbations, and lower robustness than longer ones.

Our results show that benchmark conclusions in streaming \ac{CL} are not taskification-invariant. Evaluation depends not only on the learner and the data stream, but also on how that stream is taskified. Future work should extend this analysis to adaptive and distribution-informed taskifications and investigate \ac{CL} methods that are robust to taskification variability.